# Handling uncertainty using features from pathology: opportunities in primary care data for developing high risk cancer survival methods

GOCE RISTANOSKI

Department of Computing and Information Systems, The University of Melbourne, Australia

JON EMERY

Department of General Practice and Centre for Cancer Research, Medicine, Dentistry and Health Sciences, The University of Melbourne, Australia

JAVIERA MARTINEZ GUTIERREZ

Department of General Practice and Centre for Cancer Research, Medicine, Dentistry and Health Sciences, The University of Melbourne, Australia

DAMIEN MCCARTHY

Department of General Practice and Centre for Cancer Research, Medicine, Dentistry and Health Sciences, The University of Melbourne, Australia

UWE AICKELIN

Department of Computing and Information Systems, The University of Melbourne, Australia

More than 144 000 Australians were diagnosed with cancer in 2019. The majority will first present to their GP symptomatically, even for cancer for which screening programs exist. Diagnosing cancer in primary care is challenging due to the non-specific nature of cancer symptoms and its low prevalence. Understanding the epidemiology of cancer symptoms and patterns of presentation in patient's medical history from primary care data could be important to improve earlier detection and cancer outcomes. As past medical data about a patient can be incomplete, irregular or missing, this creates additional challenges when attempting to use the patient's history for any new diagnosis. Our research aims to investigate the opportunities in a patient's pathology history available to a GP, initially focused on the results within the frequently ordered full blood count to determine relevance to a future high-risk cancer prognosis, and treatment outcome. We investigated how past pathology test results can lead to deriving features that can be used to predict cancer outcomes, with emphasis on patients at risk of not surviving the cancer within 2-year period. This initial work focuses on patients with lung cancer, although the methodology can be applied to other types of cancer and other data within the medical record. Our findings indicate that even in cases of incomplete or obscure patient history, hematological measures can be useful in generating features relevant for predicting cancer risk and survival. The results strongly indicate to add the use of pathology test data for potential high-risk cancer diagnosis, and the utilize additional pathology metrics or other primary care datasets even more for similar purposes.

# 1 INTRODUCTION

In many healthcare systems primary care clinicians are the first contact point and provide longitudinal generalist care for the whole population. Primary care often plays a key gatekeeper function to hospital specialist care; They order initial investigations and treatments and determine referral pathways should additional opinions be required. Through one's medical history, we come in contact with primary care clinicians more often than with specialists. This longitudinal nature of primary care means that primary care medical records contain a rich dataset describing individuals' medical histories, investigations, prescriptions and health outcomes. Nevertheless, the use of primary care datasets can be limited for health conditions that are difficult to diagnose or treat, have a high mortality rate or for which treatment is provided by other institutions. This provides additional challenges for any research conducted by using primary care data only.

General practitioners play a key role in early detection of serious diseases, including cancer – a condition with major impact on the patients' quality of life. The low prevalence of cancer in primary care, and the fact that many cancers present with symptoms that have more common benign causes or only appear at more advanced stages of cancer, creates significant challenges for early cancer detection. There have been significant advances in the last 15 years in our understanding of the epidemiology of cancer symptoms in primary care, arising from analyses of large general practice datasets. This has led to the creation of risk prediction models [1] and risk assessment tools based on combinations of symptoms, physical signs and test results [2]. The approaches used have a tendency not to account for trends over time of test results, nor use more complex patterns of presentations, symptoms, and test results. A GP may have several years of data available for their patients, with detailed pathology results and multiple visits over time. This creates the opportunity to investigate relationships between different pathology test results over time and cancer diagnosis. In this paper we focus our work on examining trends in the results from the frequently ordered full blood count, including platelet count, and markers of anaemia (Hb, MCV, MCH and RDW), and their association with cancer diagnosis and outcome. This in practice can lead to a faster referral to additional tests or specialist and provide an opportunity to start treatment at an earlier date.

In this work, we examine the features of pathology test results in different patient cohorts to identify patients at increased risk of an undiagnosed prevalent cancer and predict survival. We focus on predicting whether a patient will survive the 2 years period from the date the cancer diagnosis is delivered. With the patients already experiencing some symptoms when visiting the GP and getting the cancer diagnosis, we believe the prediction of the 2 years survival is a relevant decision support that will allow the GP to pursue more immediate actions for these patients. The findings of this work can be easily implemented with the current GP practices. When discovered and understood, the novel features in the pathology history for these patients will allow for a more focused investigation of the pathology history of patients for early cancer detection. Finding and generating these features from the primary care dataset is the main task researched in this paper.

We also investigate the limitations of the uncertainty of the data: irregular blood tests, infrequent tests, and tests taken months before the diagnosis date. We highlight the potential to flag patients at risk on the basis of this limited patient history. The contributions presented in this paper are as follow:

- We present clear indication of the effect of out of range (abnormal) pathology test results and patient living status for patients with lung cancer
- We associate other patient features to best determine the tasks and patient cohorts where pathology test results can be used

- We suggest methods to generate relevant features that can be used for cancer risk prediction models.
- We deliver promising results for prediction of living status for patients for 2-year period in order to determine patients at high risk
- We demonstrate the consistency of the generated features when building prediction models on subsets of the data, removing a great deal of the uncertainty factor regarding feature selection.

## 2 RELATED WORKS

Lung cancer is the most common cancer in the world. It accounted for more than 2.9 million deaths worldwide in 2018 [3]. Its high mortality is partially due to the fact that most lung cancer patients are diagnosed at late stages of their cancer given the unspecific nature of its symptoms. Hence, early detection and referral in primary care becomes crucial. Anaemia has long been recognised as a predictor of cancer [4]. Using a large primary care dataset in the UK and Scotland [1], an international study calculated an adjusted hazard ratio for lung cancer of 1.75 for females and 1.89 for males with anaemia. Thrombocytosis or raised platelet count has recently been considered as a risk predictor for several cancers worldwide. A systematic review demonstrated that thrombocytosis is a risk predictor for many cancer sites including colorectal, lung, ovary, bladder, kidney, pancreas, oesophago-gastric and uterus [5]. Using primary care data and cancer registry data in the UK, [6] studied the overall risk of cancer in patients with thrombocytosis. Lung and colorectal cancer were the most frequently diagnosed cancers in these patients. Lung cancer was diagnosed in 23% of males and 14% of females with thrombocytosis compared with 14% and 12% respectively in the general population. Probability of cancer with thrombocytosis alone at 4.2% for smokers and 1.6% for non-smokers is also investigated [2]. In Australia, both anaemia and thrombocytosis have been included in several guidelines for early cancer diagnosis.

The latest lung cancer guidelines require an urgent x-ray if patients have had a range of unexplained persistent symptoms, including thrombocytosis [7]. There have been several studies aiming to validate risk prediction tools and algorithms for lung cancer patients in primary care using primary care datasets [9][10][11]. A systematic review described the many variables of the different studies including anaemia and thrombocytosis in some cases [8]. Indications that it takes at least 3 months for symptoms to show has also been provided [12], and patients have had more frequents visits to their GP right before the initial diagnosis [14]. As some models using different blood tests have been used with AI [13][15][16][17], the use of combined metrics for cancer prediction is the next step in the research process, and it is the focus of the work presented in this paper.

## 3 MATERIALS AND METHODS

### 3.1 Dataset description

The NPS MedicineInsight initiative was established by the Australian Government Department of Health (DoH). It is a nationally representative primary care dataset from 534 general practices with more than 5000 GP providers. For this study, we used a Victorian subset of the MedicineInsight data, which forms part of the Victorian Comprehensive Cancer Centre linked datahub. This incorporates general practice data from 103 Victorian practices with approximately 1.8 million patients for the period 2007-17. It includes 8.1 million recorded diagnoses, 23 million prescriptions, 32 million encounters and 85 million pathology test results.

We identified a cohort of patients with a recorded diagnosis of lung cancer and GP ordered test results for: Platelet count, MCV (Mean Corpuscular Volume), MCH (Mean Corpuscular Hemoglobin - average mass of hemoglobin per red blood cell), MCHC (Mean Corpuscular Hemoglobin Concentration - concentration of hemoglobin in a given volume of packed red blood cell) and RDW (Red blood cell distribution width).

For each of these pathology measures there was variation by reporting laboratory of the normal range for some parameters; for example, MCV can have lower boundary from 75-80, and upper boundary 98-100 femtoliters (fL). We worked with the reported normal ranges per each individual pathology test. The pathology tests are taken at irregular intervals, with some patients having only 1-2 pathology tests, while others as high as 10, which serves as one of the uncertainty factors in the design of prediction models later on.

### 3.2 Initial cancer survival rate analysis

We begin investigating the relationship between platelets and other haematological measures to cancer by looking at the patient living statuses. Each of the different pathology measures listed have different numbers of patients due to the differences between laboratories in pathology tests conducted. Consequently, not all pathology test results will be delivered in the same format. This initial analysis aims to determine whether patients that had out of range pathology test results had different survival rates compared to patients with in-range pathology test results, and what features might be associated with these findings that can later be used for building prediction models.

As we are looking into the cases where we have a cancer diagnosis assigned, the results for platelets and haematological measures of interest recorded near the diagnosis date are of particular interest. From the date of first occurrence of the diagnosis, we consider the 3-month period of pathology tests taken from 2 months prior to 1 month after the diagnosis date to be considered as "close to the diagnosis" date, and the remaining tests taken earlier as "not close to the diagnosis" date. This assigns our patients into 3 groups:

1. Patients with no out-of-range pathology results
2. Patients with out-of-range pathology results outside the 3-month cancer diagnosis date time period
3. Patients with out-of-range pathology results within the 3-month cancer diagnosis date time period.

This type of data allows us to investigate the opportunity to group the patients in the three groups listed above solely on the occurrence of an out-of-range pathology test result, not on frequency time sequence of results. The group of patients with no out of range pathology test results is the largest patient group for each pathology measure, covering around 70-75% of the patients. The patients with out of range pathology test results, both within the 3-month period and outside of it had similar representations in terms of number of patients, with the group outside the 3-month period being more represented. Summary of the initial analysis of the patients final living status for each of these groups is presented in **Error! Reference source not found.** As we are examining the relationship between patient history and living status, we have a very strong indication that patients belonging to the second and third groups, where out-of-range pathology test results are present, exhibit higher percentages of patients that are deceased, for all of the pathology tests. Additionally, patients that belong in the group with out of range pathology measures within 3-months of cancer diagnosis have the highest portion of patients deceased for all of the pathology measures: for platelets and RDW, the difference in the percentage of deceased patients between the out of range within 3-months of cancer diagnosis date group and without out of range group is almost 15% (52.54% vs 38.48% for platelets and 49.17% vs 34.35% for RDW). This indicates we should consider the presence of an out of range pathology test as a potential feature for determining high-risk patients. This is later on investigated with not just the previously defined 3-month period, but also a 6-month period to see if we can utilize even more of the patients' pathology history.

| Measure | No out of range measure | | Out of range measure, outside of 3-months of cancer diagnosis date | | Out of range measure, within 3-months of cancer diagnosis date | |
|---|---|---|---|---|---|---|
| | Num. of patients | Percentage | Num. of patients | Percentage | Num. of patients | Percentage |
| **Platelets** | 431 | 38.48 | 93 | 46.04 | 93 | **52.54** |
| **MCV** | 405 | 41.16 | 115 | 38.59 | 67 | **51.15** |
| **MCH** | 348 | 41.09 | 135 | 40.42 | 69 | **47.26** |
| **MCHC** | 352 | 40.05 | 101 | 38.35 | 73 | **52.52** |
| **RDW** | 257 | 34.36 | 69 | 44.81 | 59 | **49.17** |

**Table 1. Patient pathology history vs. living status for deceased patients.**

This strong indication that patient's pathology history can be relevant to the living status of a patient leads to the research direction of investigating the possibility to use the pathology history for classification of patients that are potentially at risk of not surviving the cancer. With the primary care dataset not containing any specifics about the potential cancer treatment, we will investigate the more specific relationship between the pathology measures and cancer survival status by using the more standard patient features present in primary care dataset, such as age and sex.

### 3.3 Survival analysis per age group

The indication that pathology history may be relevant to the cancer patient's living status is visible in more detail when we analyse the three groups of patients per age group for Platelet test results. See ***Table 2.* Survival analysis per age group**. Patients were grouped by the first digit of their age (Group 1 for patients 10-19, etc.), and we note there is a significant difference between the group of patients without out of range pathology test results, and the group with out of range results within 3-months of diagnosis date: the lines representing the final living status are in different order based on magnitude. The most affected groups of patients are those in the age range 60-89 years old, as it is the largest group of all the patient groups.

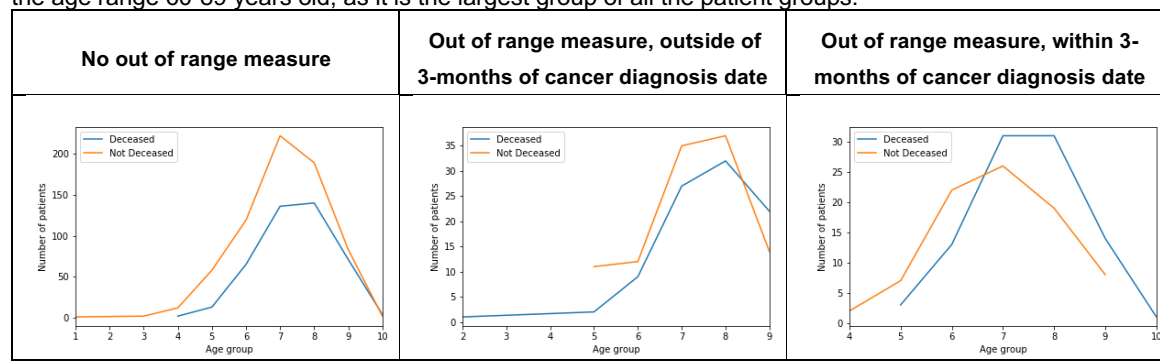

**Table 2. Survival analysis per age group for Platelets**

### 3.4 Survival analysis per biological sex

Pathology history shows strong relevance to the cancer patients living status when we group the patients per biological sex, as shown in *Table 3.* **Percentage of patients not surviving within 2 years of diagnosis date, per sex**. Patients that have out of range pathology test results have the lowest survivorship rates within 2 years from the diagnosis date, for most of the pathology measures examined. The results indicate some already known trends that male patients have a higher mortality rates than female patients [9], but this difference is amplified between the groups based on pathology history. In this case, we see an alarming indication that the out of range measures have a particular effect on female patients: based on platelets results, the mortality rate for female patients jumps from 71.25% for female patients with no out of range platelet readings, to 89.29% for female patients with out of range platelet results within 3-months of the cancer diagnosis date.

| Measure | No out of range measure | | | Out of range measure, outside of 3-months of cancer diagnosis date | | | Out of range measure, within 3-months of cancer diagnosis date | | |
|---|---|---|---|---|---|---|---|---|---|
| | Sex | Num. of patients | Percentage | Sex | Num. of patients | Percentage | Sex | Num. of patients | Percentage |
| **Platelets** | Female | 114 | 71.25 | Female | 23 | 79.31 | Female | 25 | **89.29** |
| | Male | 181 | 82.27 | Male | 45 | 84.91 | Male | 46 | **92** |
| **MCV** | Female | 101 | 72.66 | Female | 38 | 79.17 | Female | 20 | **80** |
| | Male | 179 | 84.04 | Male | 47 | **87.04** | Male | 30 | 85.71 |
| **MCH** | Female | 92 | 73.02 | Female | 38 | 79.17 | Female | 19 | **79.17** |
| | Male | 153 | 84.07 | Male | 57 | 82.61 | Male | 35 | **92.11** |
| **MCHC** | Female | 82 | **76.64** | Female | 34 | 75.56 | Female | 25 | 75.76 |
| | Male | 169 | 84.08 | Male | 37 | 84.09 | Male | 30 | **88.84** |
| **RDW** | Female | 59 | 77.63 | Female | 27 | 81.82 | Female | 18 | **85.71** |
| | Male | 117 | 81.82 | Male | 24 | 88.89 | Male | 30 | **100** |

**Table 3. Percentage of patients not surviving within 2 years of diagnosis date, per sex**

## 4 EXPERIMENTS

### 4.1 Soft range definition

For haematological measures, normal parameters are established within a range. The range is the difference between the highest and lowest normal values. As an example, the platelet count most commonly is defined within normal range of 150-450 thousand platelets per microlitre of blood. We defined soft range as being the

2.5% ends from within the lowest and highest values. For the platelets example, 2.5% of the 450-150=300 is 7.5 units, so the soft ranges would be (162.5-442.5). Even though a platelet count reading falling within these end intervals is still considered within normal range, for the purpose of our work we will investigate the labelling of platelets within these end intervals as out of range, aka, soft range. The reasoning behind this is that cases of 149 are considered out of range, but 150 is within normal range, and the relative difference between them is small. The other haematological measures have even smaller ranges, so the boundary cases just within the normal range can now be investigated as out of range, and we can see how that affects our analysis of the cancer survival rate.

### 4.2 Original features definition

With the concept of using the pathology test results to better understand the patients' cancer status and living status being new, we will investigate several combinations of labelling the five original features. The 6 versions of labelling are presented in *Table 4.* **Original features labeling versions**. For version 3 and 5, we use the time frame of 6 months prior to the cancer diagnosis date instead the 3 months (2 prior + 1 after diagnosis date).

| Version | Description | Version | Description |
| --- | --- | --- | --- |
| V1 | 1 - No out of range measure<br>1 - Out of range measure, not close to cancer diagnosis date<br>2 - Out of range Measure, close to cancer diagnosis date | V4 | 1 - No out of soft range measure<br>2 - Out of soft range measure, all dates before cancer diagnosis date<br>3 - Out of range measure, close to cancer diagnosis date |
| V2 | 1 - No out of range measure<br>2 - Out of range measure, not close to cancer diagnosis date<br>2 - Out of range measure, close to cancer diagnosis date | V5 | 1 - No out of soft range measure, irrelevant of cancer diagnosis date<br>2 - Out of soft range measure, irrelevant of cancer diagnosis date |
| V3 | 1 - No out of range measure<br>1 - Out of range measure, not close to cancer diagnosis date (more than 6 months to diagnosis date)<br>2 - Out of range measure, close to cancer diagnosis date (6 months or less to diagnosis date) | V6 | 1 - No out of soft range measure<br>2 - Soft out of range measure, not close to cancer diagnosis date (from 6 months to 2 months prior diagnosis date)<br>3 - Out of range measure, close to cancer diagnosis date |

**Table 4. Original features labeling versions**

The labelling works on the principle of a higher value labelling taking priority when a patient belongs to more than 1 group. For example, in version 1, when a patient has an out of range measure reading both close to the diagnosis date (label 2) and not close to the diagnosis date (label 1), it is label 2 that will be assigned to the patient.

### 4.3 Deriving additional combinatory features

We have the original 5 haematological features, as well the features about sex and age group. We will make 2 types of combinations:
- Combinations between the original 5 haematological measures: we will do combination of two and three features together, with additive (+) and multiplicative (*) factor. We want to investigate if a patient having both labels as the highest will have the new feature carrying more information in regard to their chances of not surviving the cancer within 2-year period.
- Combination of the newly generated features + sex: we will do additional set of each of the newly generated features, with sex as additive factor. For this purpose, the labels for sex are 1 – male, and 20 – female, to allow enough difference in values when the combinations are applied.

## 5 EXPERIMENTS

### 5.1 Dataset features

As we are looking to investigate the relationship between a patient not surviving the cancer within 2 years of the diagnosis date and the haematological measures, we will only include patients that have the original diagnosis date at least 2 years from the final date in the collected data. This will ensure that if the patient has been diagnosed with cancer, during data collection time more than 2 years have passed, so we know their status at the end of that period. We will also be looking at patients that have had pathology results for all 5 haematological measures, as using the standard techniques for dealing with missing values may not be the best approach in this case, for it may introduce a lot of noise and due to the fact that we want to fully investigate the relationship between the patient living status and the haematological measures without any assumptions for missing data.

Our total dataset has 472 patients, with a total of 149 features (original 5 haematological measures, sex, age group, total sum of original 5 haematological measures and the additional features described in 4.3).

### 5.2 Model and features selection

We decided to test the classification opportunities for our datasets using 3 models: Decision Tree, AdaBoost and XGBoost. The reason behind using Decision Tree is the interpretability of the model: we can visualise the model and try to derive additional conclusions regarding the features used. We used AdaBoost and XGBoost to allow additional increases in performance.

We test the performance of our models with 10 cross fold validation. For each of the folds, the same dataset will be used for all three models: we will generate the fold and then train the model, rather than do an individual 10 cross fold validation per model. Analysing the feature selection per model is something of interest, so we are going to be using the same dataset entirely for each model. The number of features selected will be within the range of 5-25. We will follow how the performance changes as we add more features, and what types of features top the list of most relevant. We will use the chi-squared statistic for selecting the top features per fold.

### 5.3 Accuracy

Presented in *Table 5. **Accuracy (in%) of the classification models, per number of features*** is the classification accuracy for all 6 versions. We can observe that all versions produce classification accuracy of just under 65%. This performance may seem below optimal, but it is in fact a strong indicator that pathology tests contain information about the relationship between the cancer and the survival rate. By using only 5 blood test metrics initially, and with no information about the stage of the lung cancer nor the treatment used, as well as uncertainty about the exact diagnosis period, we still were able to demonstrate the potential to use the pathology tests for classifying high-risk patients, meaning that these tests may be used for early detection of cancer tasks, provided additional features are investigated and some of the uncertainty arising from the dataset is removed. With the number of patients not surviving the 2-year period being a large portion of all cancer patients, these models have strong impact and can provide decision support even with the achieved accuracy.

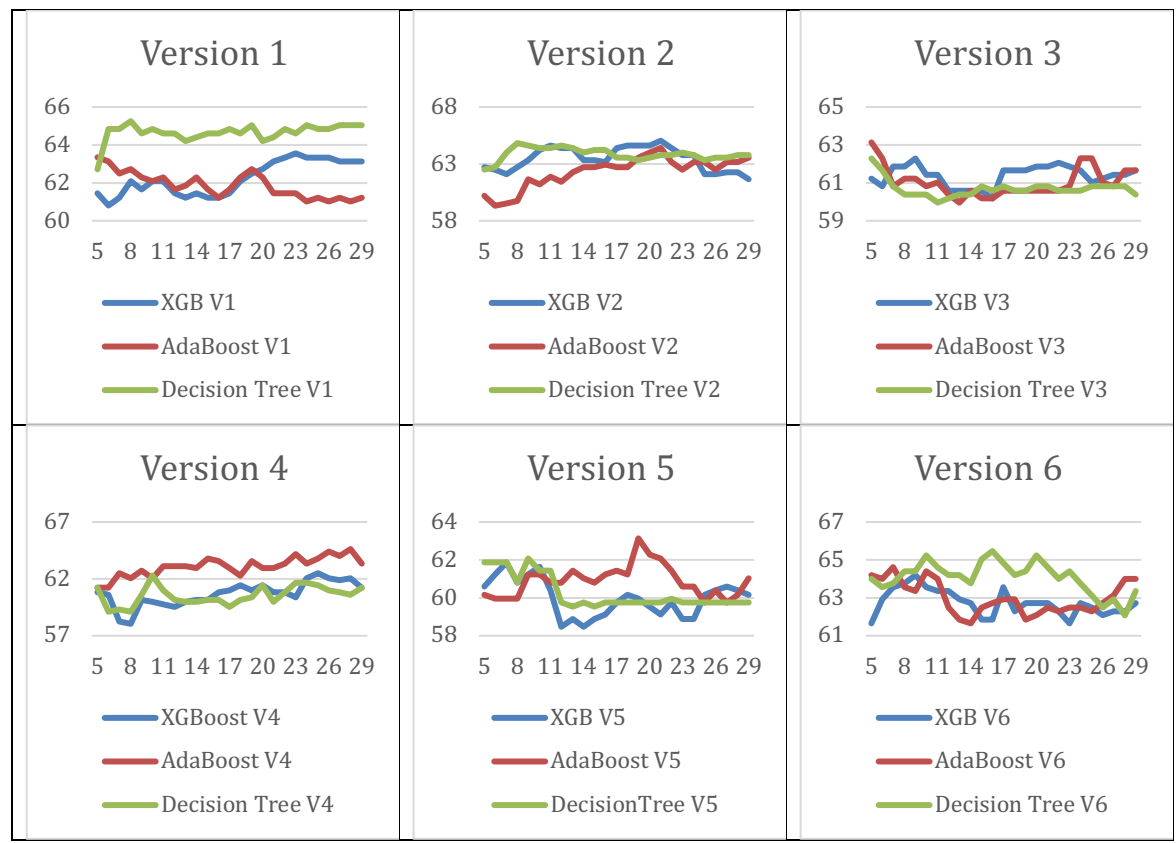

Table 5. Accuracy (in%) of the classification models, per number of features

## 6 CONCLUSION

In this paper we have investigated the use of pathology results to predict whether a patient will survive the 2 year period after their lung cancer diagnosis date. The work focused on several aspects of the primary care data, the first being the use of the right features from a full blood count test. We demonstrated how the 5 selected features used in standardized full blood count pathology tests can be indicators for the death rate of cancer patients, in particular when different patients' cohorts using age and sex are investigated. The strong uncertainty factor coming from scarce patient history regarding pathology tests was handled by deriving features from the medical record that only use occurrence of out of range pathology test results. This type of data engineering showed promising results in the design of classification models used to alert medical health professionals if the patient in question is at risk of not surviving the 2-year period after their lung cancer diagnosis. With a relatively small number of patients relative to features generated, we also demonstrated that the most relevant features are consistent when different subsets are used in the training stage. This both confirms the useful information in the pathology tests when it comes to better detecting high risk cancer patients and directs us to consider more additional features in the data in potentially different formats, to further increase the classification accuracy.

The results from this research work present several future research directions. Using a better definition of the out of range results relevant to the time stamps and other in-range results can yield more descriptive insights about the patient cohorts of highest relevance. With the number of patients having extensive pathology history being relatively small, we can attempt to derive more accurate and explanatory models for this patient cohort and support more frequent pathology tests for certain patient groups as a future practice. This will lead to the goal of using pathology tests to aid in early detection of cancer. With the results of this paper strongly indicating the relevance of the pathology tests measured, we can determine more research directions such as time series patterns of irregularity in the pathology tests for early cancer diagnosis, or other pathology features depending on the cancer type.